\documentclass[10pt,twocolumn,letterpaper]{article}

\usepackage{wacv}              

\usepackage{graphicx}
\usepackage{amsmath}
\usepackage{amssymb}
\usepackage{booktabs}
\usepackage{pifont}
\newcommand{\cmark}{\ding{51}}%
\newcommand{\xmark}{\ding{55}}%

\usepackage{colortbl}
\definecolor{Blue}{rgb}{0.6484,0.6484,0.9414}
\definecolor{Green}{rgb}{0.6484,0.9414,0.6484}

\definecolor{red-text}{rgb}{0.75390625,0.16015625,0.1796875}
\definecolor{green-text}{rgb}{0.171875,0.4296875,0.28515625}
\definecolor{blue-text}{rgb}{0.24609375,0.53125,0.76953125}

\usepackage[pagebackref,breaklinks,colorlinks]{hyperref}

\usepackage[capitalize]{cleveref}
\crefname{section}{Sec.}{Secs.}
\Crefname{section}{Section}{Sections}
\Crefname{table}{Table}{Tables}
\crefname{table}{Tab.}{Tabs.}

\def\wacvPaperID{2253} 
\def\confName{WACV}
\def\confYear{2025}

\begin{document}

\title{MoRAG - Multi-Fusion Retrieval Augmented Generation for Human Motion}

\author{
\normalsize{Sai Shashank Kalakonda}\\
\normalsize{CVIT, IIIT Hyderabad} \\
{\tt\small sai.shashank@research.iiit.ac.in} \\
\and \normalsize{Shubh Maheshwari} \\ 
\normalsize{University of California San Diego} \\
{\tt\small shmaheshwari@ucsd.edu} \\
\and \normalsize{Ravi Kiran Sarvadevabhatla} \\ 
\normalsize{CVIT, IIIT Hyderabad} \\
{\tt\small ravi.kiran@iiit.ac.in } \\
}

\twocolumn[{%
\renewcommand\twocolumn[1][]{#1}%
\maketitle

\begin{center}
  \centering
   \includegraphics[width=0.85\linewidth]{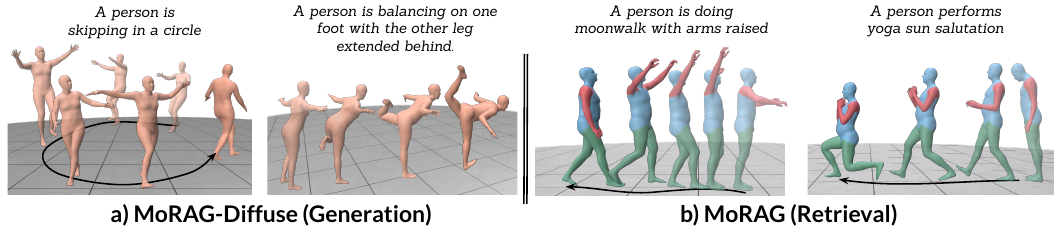}

   \captionof{figure}{\textbf{MoRAG} is a retrieval-augmented framework for generating human motion from text. It integrates part-specific motion retrieval models with large language models to improve the quality of generation and retrieval tasks across various text descriptions. The black arrow illustrates motion translation. In the bottom figures, red, blue, and green represent the retrieved motion for the \textcolor{red-text}{\textbf{hands}}, \textcolor{blue-text}{\textbf{torso}}, and \textcolor{green-text}{\textbf{legs}}. The varying transparency in the figure indicates the progression of time steps. }
   \label{fig:teaser}
\end{center}%
}]

\begin{abstract}

   We introduce \textbf{MoRAG}, a novel multi-part fusion based retrieval-augmented generation strategy for text-based human motion generation. The method enhances motion diffusion models by leveraging additional knowledge obtained through an improved motion retrieval process. By effectively prompting large language models (LLMs), we address spelling errors and rephrasing issues in motion retrieval. Our approach utilizes a multi-part retrieval strategy to improve the generalizability of motion retrieval across the language space. We create diverse samples through the spatial composition of the retrieved motions. Furthermore, by utilizing low-level, part-specific motion information, we can construct motion samples for unseen text descriptions. Our experiments demonstrate that our framework can serve as a plug-and-play module, improving the performance of motion diffusion models. Code, pre-trained models, and sample videos are available at \url{motion-rag.github.io}.
\end{abstract}

\section{Introduction}
\label{sec:intro}

Text-driven human motion generation has seen unprecedented growth in recent years\cite{zhu2023humanmotiongenerationsurvey,zhang2023remodiffuse,zhang2023finemogen,zhang2022motiondiffuse,tevet2023human, Action-GPT}. Numerous works have been proposed for this task, ranging from encoder-decoder style architectures \cite{language2pose, Ghosh_2021, Guo_2022_CVPR} to the recent emerging trend of diffusion-based models\cite{zhang2022motiondiffuse,chen2023executing,zhang2023finemogen}, which generate fine-grained, realistic motion sequences. While they can generate high-quality motion sequences for simple or familiar text descriptions similar to those in the training set, they perform poorly with complex or unseen text descriptions.

Retrieval-augmented Generation (RAG) has gained significant attention in recent years for its potential to enhance generative models by incorporating additional information through retrieval methods \cite{gao2024retrievalaugmentedgenerationlargelanguage,zhao2024retrievalaugmentedgenerationaigeneratedcontent}. By integrating retrieval-based techniques with generative models, RAG produces outputs that are more accurate, contextually relevant, and reliable. Moreover, this additional information helps enhance the model's generalizability across language space and also improves the stochasticity. However, the application of RAG in motion generation is underexplored.

A RAG system typically comprises two key components: the retriever and the generator. The retriever identifies relevant information from a database based on the input query, while the generator uses both the input query and the retrieved information to generate the desired content.

Recently proposed text-to-motion retrieval approaches\cite{petrovich23tmr,yin2024trimodal} aim to retrieve full-body motion sequences from the motion database using a contrastive training strategy between text and motion embeddings. However, these retrieval strategies do not perform well when text phrases contain spelling errors, rephrased text sequences, or substitution of synonymous words. (See Fig.~\ref{fig:llm_importance})

\begin{table}[!t]
\centering
\resizebox{0.85\linewidth}{!}
{
\begin{tabular}{ c|c|c|c|c}
Motion Retrieval Method& Text Robustness& Generalizability & Diversity& Zero-shot setting \\
\midrule

TMR\cite{petrovich23tmr} & \xmark & \xmark  & \xmark & \xmark \\
\midrule

TMR++\cite{lbensabath2024} & \cmark & \xmark & \xmark & \xmark \\
\midrule
\textbf{Ours} & \cmark  & \cmark & \cmark & \cmark \\

\bottomrule
\end{tabular}
}
\captionof{table}{Comparision of text-to-motion retrieval approaches - Text Robustness (ability to handle diverse language inputs), Generalizability (adaptation to similar yet altered data), Diversity (capacity to produce varied outputs), and Zero-Shot Setting (performance on previously unseen data).} 
\label{tab:comparison_tmr}
\end{table}

Action-GPT\cite{Action-GPT}, TMR++\cite{lbensabath2024} prompt large-language models (LLMs) to provide detailed descriptions as input. However, these approaches are limited in their ability to generate or retrieve motion sequences for text descriptions that are not present in the database, restricting the diversity of output motion sequences and reducing generalization to out-of-domain or unseen text descriptions ( Fig.~\ref{fig:quali_results} (a))

\begin{figure}[ht]
  \centering
   \includegraphics[width=0.9\linewidth]{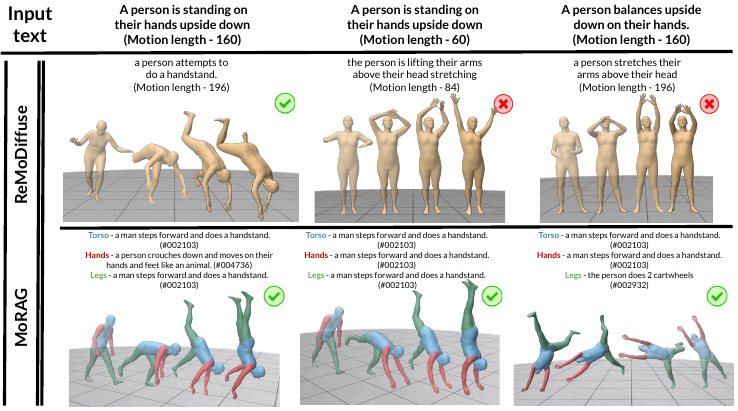}

   \caption{MoRAG utilizes part-specific descriptions to effectively retrieve relevant samples, demonstrating robustness to variations in motion length and descriptive text. In contrast, ReMoDiffuse \cite{zhang2023remodiffuse}, a hybrid approach based on motion length and text similarity, fails to retrieve suitable samples when there are changes in motion length or text. Each figure of ReMoDiffuse displays the retrieved text at the top and the corresponding motion length in brackets. For MoRAG, three part-specific retrieved texts, along with their corresponding HumanML3D ~\cite{Guo_2022_CVPR} ID, are provided using the \#. \textcolor{green}{tick} and \textcolor{red}{cross} to indicate whether the motion corresponds to the input text.}
   \label{fig:remo_retrieval}
\end{figure}

Based on the RAG concept, ReMoDiffuse\cite{zhang2023remodiffuse}, adopts a hybrid retrieval approach using motion length and CLIP\cite{clip}-based text-to-text similarity, which does not incorporate any motion-specific information, which can result in inaccurate retrievals (Fig.~\ref{fig:remo_retrieval}) and motion generation. (Fig.~\ref{fig:quali_results} (b))

To overcome these shortcomings, we propose a multi-part fusion-based augmented motion retrieval strategy that is capable of constructing diverse and reliable motion sequences. We train part-specific independent motion retrieval models that retrieve motion sequences with movements corresponding to each part aligned with the provided text description. The retrieved part-specific motion sequences are fused accordingly to construct full body motion sequences, allowing our method to even query unseen text descriptions.

Our experiments show that incorporating the constructed motion sequences as an additional conditioning to the diffusion based motion generation model improves the alignment with the semantic information of the text description and diversity of generated sequences.

In summary, our contributions are as follows: 
\begin{itemize}
    \item We propose MoRAG, a novel multi-part fusion-based retrieval augmented human motion generation framework to enhance the performance of the diffusion based motion generation model. 
    \item We adapt \textit{part-wise motion retrieval} approach that utilizes \textit{generative prompts} to construct motion sequences that align with the provided text description. 
    \item Motion sequences constructed using our retrieval strategy exhibit superior generalization and diversity, as shown by the qualitative and quantitative analysis.
\end{itemize}

\begin{figure*}[ht]
  \centering
   \includegraphics[width=0.85\linewidth]{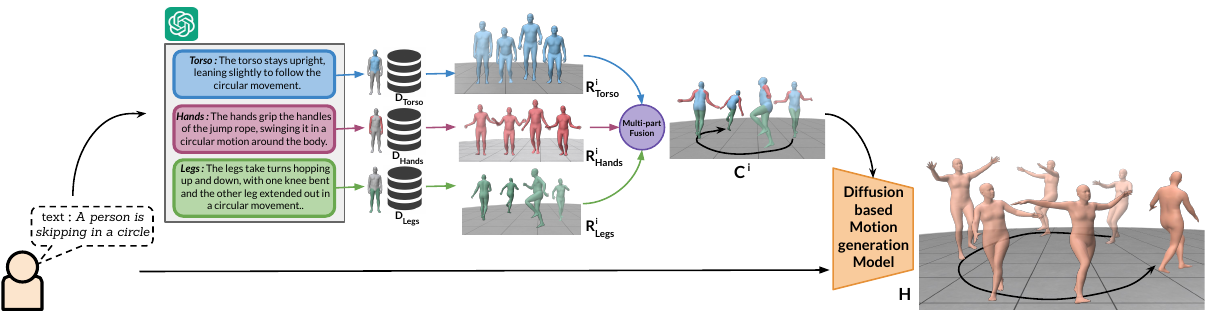}

   \caption{\textbf{MoRAG Overview}: Given a text description \texttt{text}, we generate part-specific descriptions corresponding to "Torso," "Hands," and "Legs" by prompting an LLM. These generated descriptions are used as queries to retrieve corresponding part-specific motions: $R^i_{torso}$, $R^i_{hands}$, and $R^i_{legs}$ from the motion databases $D_{torso}$, $D_{hands}$, and $D_{legs}$, respectively. The retrieved motions are then fused to construct a full-body motion sequence $C^i$ that aligns with the input text. The constructed motion samples are used as additional information in the motion generation pipeline during both training and inference, alongside the input text, to further improve model performance.}
   \label{fig:overview}
\end{figure*}

\section{Related Works}
\label{sec:related_works}

\paragraph{Text-conditioned human motion generation}
The early research efforts concentrated on encoder-decoder models with multimodal joint embedding space spanning both text and motion domains\cite{language2pose, text2action}. Text2Action\cite{text2action} proposed a GAN based generative model constituting RNN based text encoder and action decoder. JL2P\cite{language2pose} focused on learning a joint embedding space for language and pose on the motion reconstruction task using text and motion embedding. Ghosh \textit{et al.}\cite{Ghosh_2021} further improved the joint embedding space using a hierarchical two-stream model where two motion representations are learned, one for each lower body and upper body.

To enhance the text-to-motion generalizability, MotionCLIP\cite{tevet2022motionclip} incorporated image embedding of the poses into the training paradigm alongside text embedding. Both the image and text embedding are generated using CLIP.\cite{clip}. TEMOS\cite{petrovich22temos} proposed a transformer based VAE encoding approach for text and motions to improve the diversity of generated motion sequences. TEACH\cite{TEACH:3DV:2022} extends TEMOS\cite{petrovich22temos} with an additional past motion encoder to generate long motion sequences using the text description and previous sequence. T2M-GPT\cite{zhang2023generating} transforms the challenge of text-conditioned motion generation into a next-index prediction task by encoding motion sequences as discrete tokens and utilizing a transformer to predict future tokens. 

Unlike the above mentioned approaches, we propose a novel multi-part fusion-based augmented motion retrieval strategy to improve the diversity of retrieved motions and to enhance their generalizability for unseen text descriptions.

\paragraph{Motion Diffusion Models}
With recent advancements in diffusion models for text and image domain tasks, several works have been proposed in the area of text-to-motion generation. MotionDiffuse\cite{zhang2022motiondiffuse} incorporated efficient DDPM in motion generation tasks to generate diverse, variable-length, fine-grained motions. MDM\cite{tevet2023human} is a lightweight diffusion model that utilizes a transformer-encoder backbone. Instead of predicting noise, it predicts motion samples, allowing geometric losses to be used as training constraints. MLD\cite{chen2023executing} adapted the diffusion process on latent motion space instead of using raw motion sequences, generating better motions at a reduced computational overhead.

However, these diffusion-based models struggle to generalize across the language space, especially when dealing with complex or unusual text descriptions. Recent text-to-image generation works introduced retrieval-augmented pipelines in their frameworks \cite{reimagen} to address such issues. ReMoDiffuse\cite{zhang2023remodiffuse} extended MotionDiffuse\cite{zhang2022motiondiffuse} by integrating a hybrid retrieval mechanism to refine the denoising process. We improve the generalizability and diversity of ReMoDiffuse by the inclusion of large language models (LLMs) and the integration of part-specific retrieval.  

\paragraph{Text-to-Motion retrieval}

Recently, significant progress has been made by multi-modal retrieval based systems in the field of text-to-image~\cite{li2023blip2bootstrappinglanguageimagepretraining,radford2021learning,yu2022cocacontrastivecaptionersimagetext,clip} and text-to-video \cite{gao2022clip2tv,deng2023promptswitchefficientclip}. However, it has been underexplored in the field of text-to-motion due to the lack of large and diverse annotated motion capture datasets. Recent works such as BABEL\cite{BABEL:CVPR:2021} and HumanML3D\cite{Guo_2022_CVPR} have provided detailed description annotations for the large-scale motion capture collection AMASS\cite{AMASS:ICCV:2019}. Following, a few works have been proposed in the field of text-to-motion retrieval.

Initially, motion generation works\cite{Guo_2022_CVPR} used retrieval as a performance metric for evaluation purposes. TMR\cite{petrovich23tmr} is the first work to showcase text-to-motion retrieval as a standalone task. To query motions, TMR\cite{petrovich23tmr} adopted the idea of contrastive learning from CLIP\cite{clip} and extended text-to-motion generation model TEMOS\cite{petrovich22temos}. Although TMR demonstrated impressive results, there is a large scope for improvement in the generalizability of the model over language space. LAVIMO\cite{yin2024trimodal} integrated human-centric videos as an additional modality in the task of text-to-motion retrieval to effectively bridge the gap between text and motion. TMR++\cite{lbensabath2024} extended TMR by leveraging LLMs in the motion retrieval pipeline via label augmentations to increase the robustness and generalizability. However, a significant gap remains in utilizing these existing retrieval strategies for retrieval-based human motion generation approaches due to their lack of diversity and generalizability for complex or unseen text descriptions.

\section{Proposed Method}
\label{sec:proposed_method}

Fig. \ref{fig:overview} illustrates our multi-fusion retrieval-augmented human motion generation framework, \textbf{MoRAG}, which aims to enhance the performance of diffusion based motion generation model by leveraging additional motion information constructed using part-specific motion retrieval. Given an input text description \texttt{text}, we generate $N$ diverse, semantically coherent human motion sequences $\{H^1,\dots,H^n,\dots,H^N\}$. We prompt the input text description to an LLM to generate motion descriptions specific to the "Torso," "Hands," and "Legs" (Sec.\ref{sec:prompt}). These descriptions are used for the retrieval of part-specific full-body motion sequences from pre-computed part-specific motion databases (Sec.\ref{sec:mot-retrieval}). The retrieved motion sequences are then fused to construct full-body motion sequences, which serve as additional knowledge for the diffusion model (Sec.\ref{sec:mot-fusion}). This methodology enhances the model's ability to effectively handle both typical and complex/unseen input conditions (Sec.\ref{sec:morag}).

\subsection{Augmented Motion Retrieval Strategy}
\label{sec:amr}
The key component of the MoRAG framework is the retrieval of diverse and semantically aligned motion samples from the database based on the given input text query. Existing text-to-motion retrieval methods \cite{petrovich23tmr,yin2024trimodal,lbensabath2024} typically retrieve full-body motion samples directly from the database. However, these approaches overlook the fact that actions are frequently characterized by localized dynamics, often involving only small subsets of joint groups, such as the hands (e.g., ‘eating’) or legs (e.g., ‘sitting’).~\cite{trivedi2022psumnet,DSAG} This results in two significant issues: (i) limited generalizability and (ii) lack of diversity in the retrieved samples. (See Table \ref{tab:comparison_tmr} and Fig. \ref{fig:quali_results} (a)) This is due to the limited availability of text-motion annotated datasets. Although BABEL \cite{BABEL:CVPR:2021} and HumanML3D \cite{Guo_2022_CVPR} provide detailed text annotations for the large-scale motion capture collection AMASS \cite{AMASS:ICCV:2019}, they are still insufficient to generalize across the language space. However, the AMASS dataset contains extensive low-level body parts information that holds the potential to generalize across a significantly broader language space. 

Based on this observation, we design independent part-specific motion retrieval models that can retrieve full-body motion sequences with movements corresponding to specific parts aligned with the provided text description. This enables dedicated part descriptions for retrieval of actions involving specific part movement. By composing these motion samples, we can construct full-body motion sequences that are semantically coherent with the given text input.
The composition also improves the expressivity of motions, since fine-grained motion details are often expressed in text in terms of body parts. The wide variety of composing combinations provides huge diversity in the constructed motion samples. Integrating these samples into a motion generation pipeline as additional information can enhance the model's performance. We also observe better generations for unseen text descriptions. (See Fig. \ref{fig:quali_results} (b))

The objective is to construct a series of motion sequences \{$C^1,\dots,C^i,\dots,C^k$\} from the motion database ranked from $1$ to $k$ where each motion sequence $C^i$ is represented as a sequence of human poses \{$C^i_1,\dots,C^i_t,\dots,C^i_{f_i}$\} with $f_i$ representing the number of timesteps for motion $C^i$. The motion retrieval strategy in MoRAG comprises three steps: (1) generating part-specific body movement descriptions, (2) retrieving part-specific motion sequences, and (3) composing the retrieved motion sequences. Details of each are provided in the following section. 

\subsection{Generation of part-specific descriptions}
\label{sec:prompt}
Given the text description \texttt{text}, we generate part specific body movement descriptions using an LLM as a knowledge engine. We construct a suitable prompt \texttt{text$_{prompt}$},  for \texttt{text} using a prompt function $f_{prompt}$, comprising of three components: 

(i) \textbf{Task instructions}, to specify the details of our task: 
\begin{center}
"\textit{The instructions for this task is to describe the listed body parts' position and movements in a sentence using simple language. ['Torso',' Hands', 'Legs']}" 
\end{center}

(ii) \textbf{Few-shot examples}, provides a set of examples consisting of diverse action descriptions to determine the format of the output we are expecting. 

(iii) \textbf{Query}, to incorporate the input text \texttt{text} to generate part-specific body movement and orientation information. 

\begin{center}
    "\textit{Query: Describe the below body parts position and movements involved in the action [\texttt{text}] in a sentence using simple language. 1) Torso 2) Hands 3) Legs}".
\end{center}

Specifically prompting for the position provides the global orientation of the body parts which results in better retrieval. Further details along with rendered video showcasing significance of position information can be found in project page.  

\begin{figure*}[ht]
  \centering
   \includegraphics[width=0.85\linewidth]{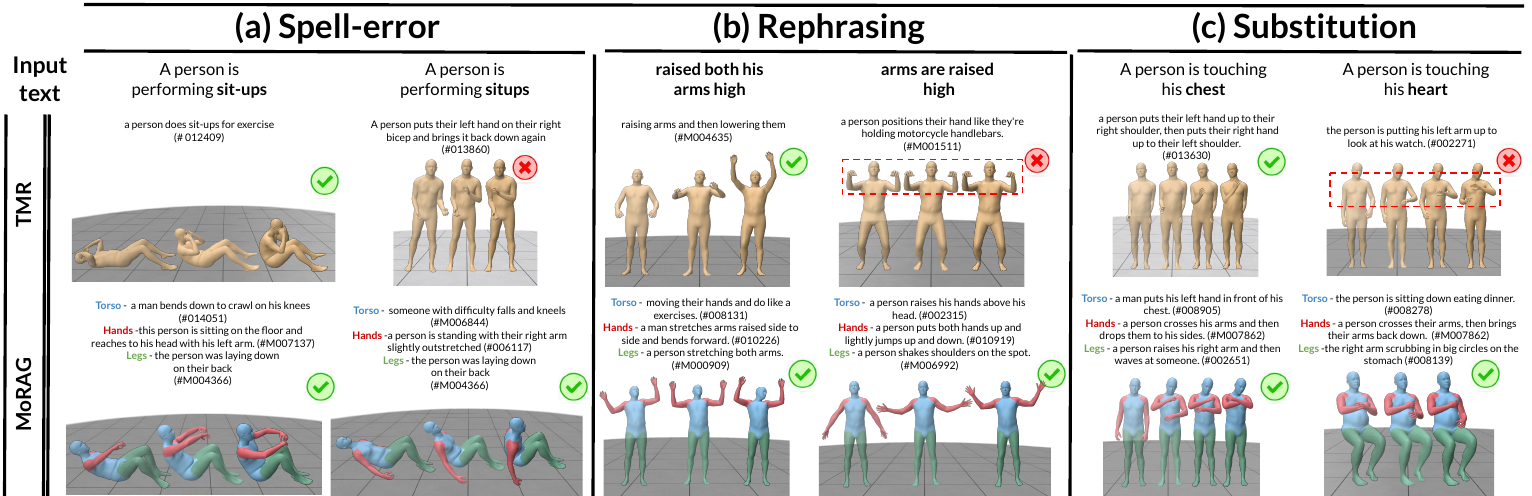}

   \caption{\textbf{LLM Importance}: Incorporating part-wise descriptions generated by LLMs into text-to-motion retrieval improves generalization over the language space. (a) \textbf{Spell Error} - MoRAG successfully retrieves and constructs the correct motion sequence when `sit-ups` is replaced with `situps`, unlike TMR\cite{petrovich23tmr}. (b) \textbf{Rephrasing} - MoRAG effectively retrieves the correct motion sequence even when the voice is changed from active to passive. (c) \textbf{Substitution} - MoRAG accurately retrieves the correct motion sequence when `chest` is replaced with its synonym `heart`.}
   \label{fig:llm_importance}
\end{figure*}

The constructed prompt is then passed through LLM to generate descriptions of the positions and movements for the specified body parts denoted as \texttt{text}$_{torso}$, \texttt{text}$_{hands}$ and \texttt{text}$_{legs}$. Training a retrieval model on these body part descriptions enables the retrieval of motions for rephrased and spell-error text phrases, thereby having a better generalization over language space. (Fig.\ref{fig:llm_importance})

\begin{figure*}[ht]
  \centering
   \includegraphics[width=0.85\linewidth]{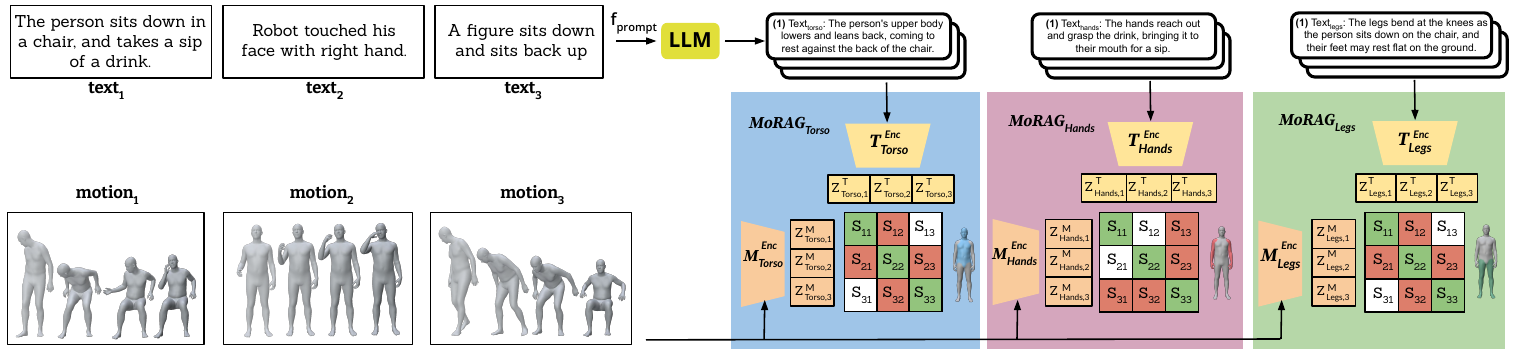}

   \caption{\textbf{MoRAG Training}: Our objective is to construct three independent part-specific motion databases. The training paradigm includes three motion retrieval models: $MoRAG_{torso}$, $MoRAG_{hands}$, and $MoRAG_{legs}$, each corresponding to a specific body part. We train these three models independently using part-specific body movement descriptions generated by LLMs for text phrases $\texttt{text}_i$ and their corresponding full-body motion sequences $\texttt{motion}_i$. We adopt a contrastive training objective between part-specific text embeddings ($Z^T_{p, i}$) generated by text encoders ($T^{Enc}_p$) and motion embeddings ($Z^M_{p, i}$) generated by the corresponding part-specific motion encoder($M^{Enc}_p$). The diagonal elements, representing positive pairs (green), are maximized, while the off-diagonal elements, representing negative pairs with text similarity below a threshold (red), are minimized. For simplicity, we do not visualize the motion decoder, but we follow a similar training procedure as described in \cite{petrovich23tmr}.}
   \label{fig:amr}
\end{figure*}

\subsection{Multi-part motion retrieval}
\label{sec:mot-retrieval}

As shown in Fig. \ref{fig:amr}, MoRAG uses $3$ independently trained TMR\cite{petrovich23tmr} models,
$ MoRAG = \{ MoRAG_{torso}, MoRAG_{hands} , MoRAG_{legs} \} $ corresponding to the respective body part. We do not train separate left and right body parts models to avoid asynchronous movements in the composed motion. Rendered video showcasing the shortcomings of seperate left and right body parts retrieval can be found in project page.
For a part $p \in \{ torso, hands, legs\}$, the retrieval model $MoRAG_p = \{T^{Enc}_p, M^{Enc}_p, M^{Dec}_p, \mathcal{D}^{Enc}_p \}$ consists of a text encoder, motion encoder, motion decoder, and motion database respectively. The model architecture for the encoder and decoder are based on TEMOS\cite{petrovich22temos}.

\textbf{Text Encoders ($T^{Enc}_p$)} : 
The LLM-generated part-specific motion descriptions of the text sequence \texttt{text} are first passed through a pre-trained and frozen DistilBERT\cite{sanh2020distilbertdistilledversionbert} to generate features $\mathcal{F}^T_{p}$ for each part-wise text description \texttt{text}$_{p}$. Along with the features $\mathcal{F}^T_{p}$, two learnable distribution tokens are passed as input to the text encoders. The outputs corresponding to the distribution tokens passed are considered as Gaussian distribution parameters $(\mu^T_p$ and $\sigma^T_p)$ from which the latent vector $Z^T_p$ is sampled using reparametrization trick\cite{VAE}.
\begin{gather}
    (\mu^T_p, \sigma^T_p) = T^{Enc}_{p} (\mathcal{F}^T_{p}) \\
    Z^T_{p} \sim  \mathcal{N}(\mu^T, \sigma^T)  
\end{gather}

\textbf{Motion Encoders ($M^{Enc}_p$)} : 
Similarly, $Z^M_p$ is obtained from the motion encoders by inputting the corresponding full-body motion sequence $M_{1:f}$ associated with the text description \texttt{text} and duration $f$. 
\begin{gather}
    (\mu^M_p, \sigma^M_p) = M^{Enc}_{p} (M_{1:f}) \\
    Z^M_{p} \sim  \mathcal{N}(\mu^M, \sigma^M)  
\end{gather}

During retrieval, instead of sampling, we directly use the embedding corresponding to the mean parameter (i.e $Z^T_p=\mu^T_p$ and $Z^M_p=\mu^M_p$)

To enhance the effectiveness of motion retrieval, we train all three motion encoders ($M^{Enc}_p$) using full-body motion sequences instead of just the respective body parts. This approach is based on the observation that LLM-generated part-specific descriptions contain information about the queried body part about other body parts. Utilizing full-body motion sequences allows us to leverage intra-joint information, resulting in more coherent and semantically accurate motion retrieval. LLM-generated part-specific descriptions for sample text inputs alongside their corresponding retrieval results can be found in project page. 

\textbf{Motion Decoders ($M^{Dec}_p$)} : 
Motion decoders input a latent vector $Z$ and sinusoidal positional encoding of the duration $f$ and output a full body motion sequence $\hat{M}_{1:f}$ non-autoregressively. The input latent vector $Z$ is obtained from one of the two encoders during training. However, since our task is motion retrieval, the decoder is not used during inference.
\begin{gather}
    \hat{M}_{1:f} = M^{Dec}_{p} (Z) \\
    Z \in \{Z^T_{p}, Z^M_{p}\}   
\end{gather}

\textbf{Loss} :
\label{tmr_loss} Each retrieval model, $MoRAG_p$ is trained with the loss $\mathcal{L}^p$ \cite{petrovich23tmr}: 
\begin{equation}
    \mathcal{L}^p =\mathcal{L}_{R}+\lambda_{KL}\mathcal{L}_{KL}+\lambda_{E}\mathcal{L}_{E} +\lambda_{NCE}\mathcal{L}_{NCE}    
\end{equation}

$\mathcal{L}_{R}$ is the motion reconstruction loss given motion and text embeddings to the decoder. $\mathcal{L}_{KL}$ is the Kullback-Leibler(KL) divergence loss composed of four losses. The first two are for the text and motion distributions with normal distribution and the other two are between text and motion distributions. $\mathcal{L}_{E}$ is the cross-modal embedding similarity loss between both text and motion latent embeddings $Z^T_p$ and $Z^M_p$. 

$\mathcal{L}_{NCE}$ is the contrastive loss which is based on InfoNCE\cite{infonce} formulation, used to better structure the cross-model latent space. The text and its corresponding motion embedding are considered positive pairs ($Z^T_{p, i}$ and $Z^M_{p, i}$), whereas all other combinations are considered to be negative ($Z^T_{p, i}$ and $Z^M_{p,j}$) with $i \neq j$.  Similarity matrix $S$ computes the pairwise cosine similarities for all the pairs, $S_{ij}=$cos$(Z^T_{p,i},Z^M_{p,j})$. However, not all negative pairs are involved in the loss computation. Text-motion pairs with text description similarities above a certain threshold, referred to as 'wrong negatives', are filtered out from the loss computation. The threshold to filter negatives is set to 0.8. These text similarities are computed using MPNet\cite{mpnet}.

\begin{equation}
    \mathcal{L}_{NCE} = -\dfrac{1}{2N}\sum_i\bigg(log\dfrac{e^{S_{ii}/\tau}}{\sum_{j}e^{S_{ij}/\tau}} + log\dfrac{e^{S_{ii}/\tau}}{\sum_{j}e^{S_{ji}/\tau}}\bigg) \\
\end{equation}

\textbf{Motion database ($\mathcal{D}_p)$ } : 
Post training, we create a database consisting of three key-value tables for every body part where each key is a unique identifier for a motion sample from the AMASS\cite{AMASS:ICCV:2019} database. The corresponding value is a vector inferred from the motion encoder. During retrieval, the LLM-generated part description is encoded into a query vector for every text encoder,  $T^{Enc}_p$. We use this query vector to search the corresponding vector indexes, finding the $k$-nearest neighbors in the embedding space using cosine similarity. The corresponding $k$ full-body motion sequences \{$R^1_p,\dots,R^i_p,\dots,R^k_p$\} are retrieved for each body part $p$.

\subsection{Spatial motion composition}
\label{sec:mot-fusion}

The retrieved motion sequences \{$R^1_p,\dots, R^i_p,\dots, R^k_p$\} are composed such that the $i_{th}$ sequence corresponding to each part $p$ is used to construct the $i_{th}$ full-body motion sequence $C^i$. This results in $k$ full-body motion sequences \{$C^1,\dots, C^i,\dots, C^k$\}, which are used as additional guidance for the motion diffusion model. To generate these top-$k$ sequences, we employed a rank-by-rank combination approach, where the retrieved part-specific sequences are selected and combined in order of their rank across different models. However, alternative combination methods could be employed to create a significantly larger number of sequences. Our composition approach is similar to SINC\cite{SINC} but we do not require the use of an LLM for mapping joints from the retrieved sequences to the composed sequence. 

To construct the composed motion sequence $C^i$ using the corresponding retrieved full-body motion sequences, \{$R^i_{torso}, R^i_{hands}, R^i_{legs}$\}, we follow these steps: (1) Trimming all three retrieved sequences to the length of the shortest one; $f_{min} = \min_{p} (f_p),$ (2) Selecting the respective body part's joint information from the corresponding retrieved motions. We follow the SMPL\cite{SMPL} skeleton structure with the first $22$ joints and partition it into three disjoint sets of joints: $J = \{j_{torso} \cup j_{hands} \cup j_{legs}\}$.
\begin{gather}
    C^i_f[j_{p}] = R^i_{p,f}[j_{p}]
\end{gather}
\begin{center}
    $p \in \{ torso, hands, legs\}, f \in [1,f_{min}]$
\end{center}
(3) Choosing the global orientation and translation from $R^i_{legs}$, as leg motion is closely associated with changes in global translation and orientation. 

\begin{figure*}[ht]
  \centering
   \includegraphics[width=0.88\linewidth]{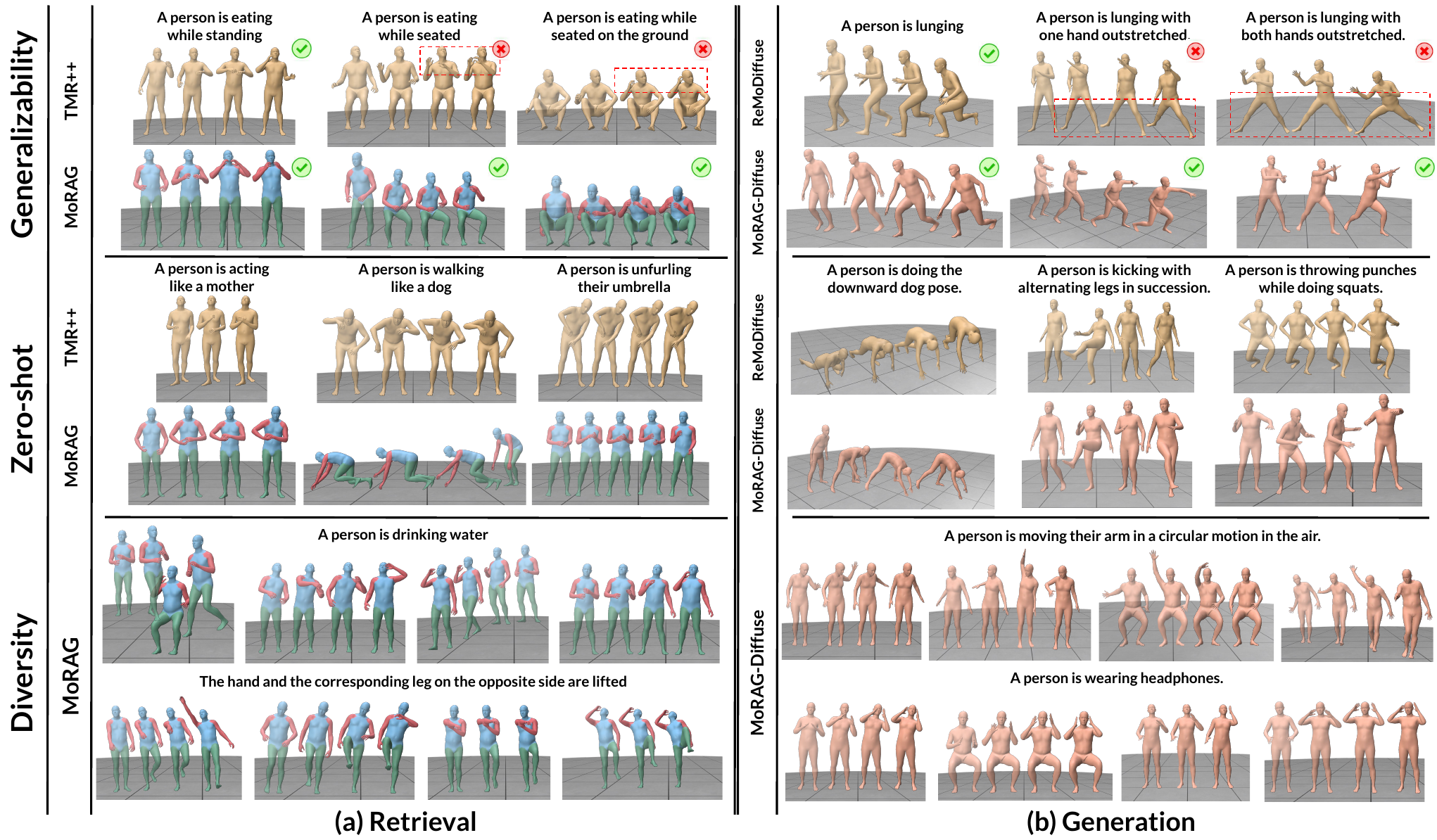}

   \caption{\textbf{Qualitative Results}: Comparison of motion retrieval and generation using our multi-part fusion approach: retrieval is compared with TMR++ \cite{lbensabath2024}, a state-of-the-art motion retrieval method and generation is compared with ReMoDiffuse \cite{zhang2023remodiffuse}. \textbf{Top}: Our method demonstrates superior generalization capabilities. \textbf{Middle}: Our approach generates accurate motion sequences for unseen text descriptions. \textbf{Bottom}: Our setup exhibits increased diversity.}
   \label{fig:quali_results}
\end{figure*}

\begin{table*}[!t]
\centering
\resizebox{0.68\linewidth}{!}
{
\begin{tabular}{c|ccc|c|c|c|c}
Methods & \multicolumn{3}{c}{R Precision $\uparrow$} & FID $\downarrow$ & MM Dist $\downarrow$ & Diversity $\rightarrow$ & MultiModality $\uparrow$\\
\midrule
 & Top 1 & Top 2 & Top 3 &  &  &\\

\midrule

Real motions & $\mathbf{0.511^{\pm 0.003}}$ & $\mathbf{0.703^{\pm 0.003}}$ & $\mathbf{0.797^{\pm 0.002}}$ & $\mathbf{0.002^{\pm 0.000}}$ & $\mathbf{2.974^{\pm 0.008}}$ & $\mathbf{9.503^{\pm 0.065}}$ & - \\
\midrule
MDM\cite{tevet2023human} & $\mathbf{0.320^{\pm 0.005}}$ & $\mathbf{0.498^{\pm 0.004}}$ & $\mathbf{0.611^{\pm 0.007}}$ & $\mathbf{0.544^{\pm 0.044}}$ & $\mathbf{5.566^{\pm 0.027}}$ & \cellcolor{Blue}{$\mathbf{9.559^{\pm 0.086}}$} &\cellcolor{Green}{$\mathbf{2.799^{\pm 0.72}}$} \\
MotionDiffuse\cite{zhang2022motiondiffuse} & $\mathbf{0.491^{\pm 0.001}}$ & $\mathbf{0.681^{\pm 0.001}}$ & $\mathbf{0.782^{\pm 0.001}}$ & $\mathbf{0.630^{\pm 0.001}}$ & $\mathbf{3.113^{\pm 0.001}}$ & $\mathbf{9.410^{\pm 0.049}}$ & $\mathbf{1.553^{\pm 0.042}}$ \\
MLD\cite{chen2023executing} & $\mathbf{0.481^{\pm 0.003}}$ & $\mathbf{0.673^{\pm 0.003}}$ & $\mathbf{0.772^{\pm 0.002}}$ & $\mathbf{0.473^{\pm 0.013}}$ & $\mathbf{3.196^{\pm 0.010}}$ & $\mathbf{9.724^{\pm 0.082}}$ & $\mathbf{2.413^{\pm 0.079}}$ \\
ReMoDiffuse\cite{zhang2023remodiffuse} & \cellcolor{Blue}{$\mathbf{0.510^{\pm 0.005}}$} &\cellcolor{Blue}{$\mathbf{0.698^{\pm 0.006}}$} &\cellcolor{Green}{$\mathbf{0.795^{\pm 0.004}}$} &\cellcolor{Green}{$\mathbf{0.103^{\pm 0.004}}$} &\cellcolor{Blue}{$\mathbf{2.974^{\pm 0.016}}$} & $\mathbf{9.018^{\pm 0.075}}$ & $\mathbf{1.795^{\pm 0.043}}$ \\
FineMoGen\cite{zhang2023finemogen} & $\mathbf{0.504^{\pm 0.002}}$ & $\mathbf{0.690^{\pm 0.002}}$ & $\mathbf{0.784^{\pm 0.002}}$ &\cellcolor{Blue}{$\mathbf{0.151^{\pm 0.008}}$} & $\mathbf{2.998^{\pm 0.008}}$ & $\mathbf{9.263^{\pm 0.094}}$ &
$\mathbf{2.696^{\pm 0.079}}$ \\

\bottomrule
\textbf{MoRAG-Diffuse} &\cellcolor{Green}{$\mathbf{0.511^{\pm 0.003}}$} &\cellcolor{Green}{$\mathbf{0.699^{\pm 0.003}}$} &\cellcolor{Blue}{$\mathbf{0.792^{\pm 0.002}}$} & $\mathbf{0.270^{\pm 0.010}}$ &\cellcolor{Green}{$\mathbf{2.950^{\pm 0.012}}$} &\cellcolor{Green}{$\mathbf{9.536^{\pm 0.104}}$} &\cellcolor{Blue}{$\mathbf{2.773^{\pm 0.114}}$} \\
\bottomrule
\end{tabular}
}
\captionof{table}{\textbf{Quantitative Results:} We compare the results of text-to-motion generation between ours and the state-of-the-art diffusion based methods on HumanML3D\cite{Guo_2022_CVPR} dataset. Our method achieves better semantic relevance, diversity, and multimodality performances. \colorbox{Green}{Indicate best results} , \colorbox{Blue}{indicates second best results.}} 
\label{tab:metrics}
\end{table*}

\subsection{MoRAG-Diffuse}
\label{sec:morag}

For generation, we extend ReMoDiffuse\cite{zhang2023remodiffuse}, a diffusion-based model, by incorporating our retrieval mechanism within the motion generation pipeline. Unlike ReMoDiffuse\cite{zhang2023remodiffuse}, we adapt our multi-part composed motion, rather than their motion length and text based similarity retrieval approach.

For the top-$k$ retrieved motion sequence $C^i$, we follow the 263 dimension motion representation as in \cite{Guo_2022_CVPR}, where each human pose $C^i_t$ is represented by $(\dot{r}^a,\dot{r}^x,\dot{r}^z,r^y,\textbf{j}^p,\textbf{j}^v,\textbf{j}^r,\textbf{c}^f)$, where $\dot{r}^a \in \mathbb{R}$ is root angular velocity along Y-axis; $\dot{r}^x \in \mathbb{R}$, $\dot{r}^z \in \mathbb{R}$ are global root velocities in X-Z plane; $\dot{r}^y \in \mathbb{R}$ is root height; $\textbf{j}^p \in \mathbb{R}^{3 \times n(J)}$, $\textbf{j}^v \in \mathbb{R}^{3 \times n(J)}$, $\textbf{j}^r \in \mathbb{R}^{6 \times n(J)}$ are the local pose positions, velocity and rotation respectively. $\textbf{c}^f \in \mathbb{R}^{4}$ is the foot contact features calculated by the heel and toe joint velocity. $n(J)$ is the number of joints and $T_i$ represents the number of timesteps for motion $C^i$.

To effectively utilize information from retrieved motion samples, we use the Semantics-Modulated Transformer (SMT) introduced in ReMoDiffuse\cite{zhang2023remodiffuse}. It comprises of $N$ identical decoder layers, each featuring a Semantics-Modulated Attention (SMA) layer and a feed-forward network (FFN) layer. The SMA layer integrates information from the input description and the retrieved samples, refining the noised motion sequence throughout the denoising process. The SMA layer consists of a cross-attention mechanism where the noised motion sequence serves as the query vector $Q$. The key vector $K$ and value vector $V$ are derived from three sources of data: (1) the noised motion sequence itself; (2) CLIP's text features of the input description \texttt{text} which are further processed by two learnable transformer encoder layers; and (3) the retrieved motion and text features $R^m, R^t$, extracted using transformer-based encoders from the retrieved samples.  As our composed motion samples do not have associated text sequences for generating text features $R^t$, we use the features of the input text description \texttt{text}.

\section{Experiments \& Results}
\label{sec:qualitative}


\subsection{Dataset and Implementation Details}
\label{sec:dataset}
We chose HumanML3D\cite{Guo_2022_CVPR} to evaluate our framework due to its extensive and diverse collection of text-annotated motions. It provides annotations for the motions in the AMASS\cite{AMASS:ICCV:2019} and HumanAct12\cite{guo2020action2motion} datasets.
We follow the same splits as provided in TMR\cite{petrovich23tmr} and ReMoDiffuse\cite{zhang2023remodiffuse} to train the retrieval and generation models respectively.

\label{sec:implementation}
We use OpenAI’s GPT-3.5-turbo-instruct for its efficiency in executing specific instructions and providing direct answers. Our prompting strategy allows a maximum of 256 tokens for both the prompt and the generation. The completions API, with default parameters, is used to generate the desired part-specific descriptions.



For part-specific retrieval models ($MoRAG_p$), a batch size of 32 is used, with other hyperparameters following the configuration in TMR \cite{petrovich23tmr}. The MoRAG-Diffuse model adopts settings similar to those of ReMoDiffuse \cite{zhang2023remodiffuse} for HumanML3D \cite{Guo_2022_CVPR}. It is trained on an NVIDIA GeForce RTX 2080 Ti with a batch size of 64, leveraging pre-trained weights from ReMoDiffuse \cite{zhang2023remodiffuse} for 50k steps.

\subsection{Results}
\label{sec:results}
Fig.~\ref{fig:quali_results} presents qualitative comparisons of MoRAG for both retrieval and generation tasks. We compare our retrieval results with TMR++\cite{lbensabath2024}, a state-of-the-art motion retrieval model, and our generation results with ReMoDiffuse\cite{zhang2023remodiffuse}, focusing on generalizability, zero-shot performance, and diversity. 

We observe that MoRAG constructed samples, utilizing part-specific retrieved motions, exhibit superior generalizability across the language space and effectively adapt to low-level changes. The richer and dedicated part-specific descriptions generated by the LLM helped retrieve precise part-specific motion sequences corresponding to the input text. Multi-part fusion has improved the construction of motion samples for unseen text descriptions, achieving better semantic alignment with the input text. Current motion retrieval methods treat the input skeleton in a monolithic manner, processing all joints in the pose tree as a whole. However, these approaches overlook the significance of sub-parts, which could enhance generalizability for unseen text descriptions. By dividing each action into sub-actions corresponding to specific subsets of joints, retrieval from the database can be made more effective. For example, in Fig.~\ref{fig:quali_results} (a), the text phrase, \textit{"A person is eating while seated on the ground"}, doesn't exist in the database which leads to the retrieval of the closest matched sample by TMR++\cite{lbensabath2024}, \textit{"this person is sitting on the floor and reaches to his head with his right arm."}. However, MoRAG searches for the part-specific sub-actions "eating" and "sitting on the ground," which can be easily retrieved from the database. When composed, these sub-actions construct a relevant motion sequence. Additionally, the extensive range of composition combinations contributes to significant diversity in the constructed sequences.  

Similarly, MoRAG-Diffuse demonstrates improved generalization to the language space and generation of motion sequences for unseen text conditions, leveraging low-level information captured through retrieval conditioning. 

Table. \ref{tab:metrics} summarizes the results of using our framework in comparison with the existing diffusion-based motion generation models. Incorporating part-specific motion retrieval models as additional knowledge in the motion generation pipeline shows an improvement over Diversity, Multi-Modal Distance, and MultiModality metrics. As observed in MUGL\cite{MUGL}, quality scores such as FID based on feature representations often fail to capture the key action dynamics of the motion sequences.  We empirically observed that these scores correlate poorly with the visual quality of motion generations. 




\section{Discussion}

\textbf{Challenges in Motion Retrieval Metrics:} To evaluate text-to-motion retrieval methods such as TMR\cite{petrovich23tmr} and TMR++\cite{lbensabath2024}, the similarity between the text corresponding to the retrieved motion sample and the input text is computed. However, for our approach, which involves the spatial composition of motion sequences, there is no single corresponding text for the entire composed sequence. As a result, these metrics cannot be computed in our case.

\textbf{Assumptions and Limitations:} \textbf{(1)}\textit{GPT Dependency:} Our approach relies on GPT-generated descriptions for motion retrieval, with accuracy dependent on the precision of these descriptions. Semantic deviations may impact retrieval quality. Feedback mechanisms or quality checks can enhance robustness and mitigate inconsistencies.
\textbf{(2)}\textit{Limited Dataset:} A constrained dataset size may lead to suboptimal retrievals, particularly for complexinputs. Expanding the dataset or integrating external data can improve retrieval precision and diversity. \textbf{(3)}\textit{Composition Alignment:} Trimming retrieved motions to the shortest sequence for alignment can introduce minor semantic or temporal inconsistencies. Incorporating a coherence check can ensure better fidelity and alignment with the input text.

\textbf{Future scope:} Future work could extend our approach to other architecture-based generative models. Incorporating
additional body part information, such as finger, head, and lip movements from respective part-specific databases, would enhance the realism of generated samples and better handle more complex language descriptions. 


\section{Conclusion}
In this paper, we enhance the performance of the diffusion-based motion generation model using a multi-part fusion-based retrieval augmented motion generation framework, MoRAG. Our method incorporates additional guidance into the motion generation pipeline using the diverse motion sequences constructed from the samples retrieved from part-specific motion databases. We propose a simple solution to construct multiple diverse, semantically coherent motion samples from the database, even for unseen descriptions, which is not feasible with existing full-body motion retrieval approaches. This makes MoRAG a more viable alternative for text-to-human motion generation by combining the strengths of both retrieval models — rapid motion sample construction and generative models — the ability to create novel outputs.



\noindent \textbf{\LARGE Appendix}


\begin{figure*}[!t]
  \centering
   \includegraphics[width=0.85\linewidth]{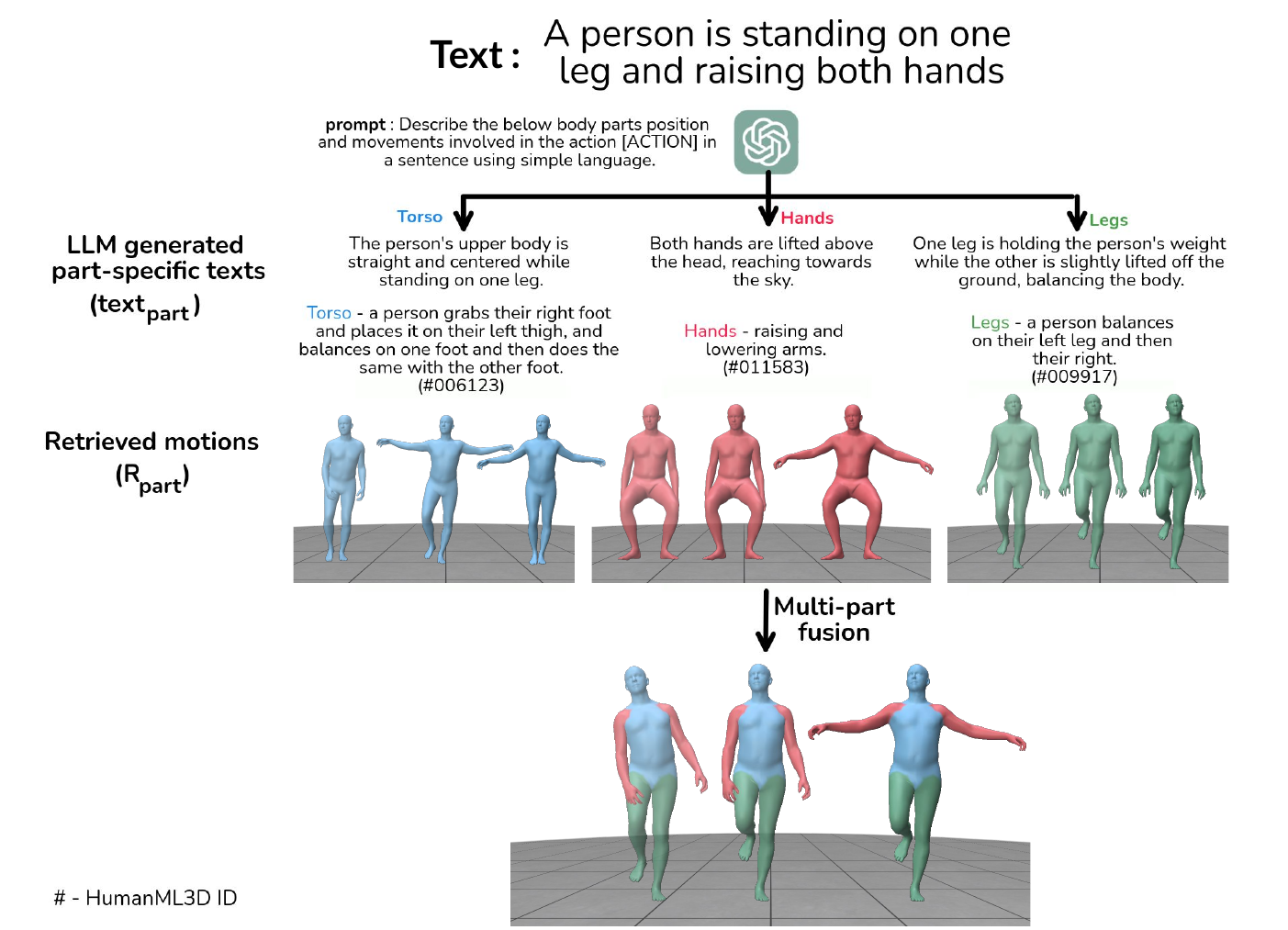}

   \caption{\textbf{Spatial Motion Composition}: Illustration of spatial motion composition using retrieved part-specific motion samples for the text: \textit{"A person is standing on one leg and raising both hands"}}
   \label{fig:spatial_comp}
\end{figure*}

\section{Dataset}
\label{sec:dataset}
We chose HumanML3D\cite{Guo_2022_CVPR} to evaluate our framework due to its extensive and diverse collection of motions paired with a wide range of text annotations. It provides annotations for the motions in the AMASS\cite{AMASS:ICCV:2019} and HumanAct12\cite{guo2020action2motion} datasets. On average, each motion is annotated three times with different texts, and each annotation contains approximately 12 words. Overall, HumanML3D consists of 14,616 motions and 44,970 descriptions. The data is augmented by mirroring left and right. We follow the same splits as TMR\cite{petrovich23tmr} and ReMoDiffuse\cite{zhang2023remodiffuse} to train the retrieval and generation models respectively.

\section{Implementation Details}
\label{sec:implementation}
We use OpenAI’s GPT-3.5-turbo-instruct for its efficiency in executing specific instructions and providing direct answers. Our prompting strategy allows a maximum of 256 tokens for both the prompt and the generation. The completions API, with default parameters, is used to generate the desired part-specific descriptions.

For part-specific retrieval models ($MoRAG_p$), we use AdamW \cite{loshchilov2019decoupledweightdecayregularization} optimizer with a learning rate of 0.0001 and a batch size of 32. The latent dimensionality of the embeddings is 256. We set the temperature $\tau$ to 0.1, and the weight of the contrastive loss term $\lambda_{NCE}$ to 0.1. Other hyperparameter values are used similarly to those in TMR \cite{petrovich23tmr}.

For MoRAG-Diffuse, we use similar settings as that of ReMoDiffuse\cite{zhang2023remodiffuse} used for HumanML3D\cite{Guo_2022_CVPR}. For the diffusion model, the variances $\beta_t$ are spread linearly from 0.0001 to 0.02, and the total number of diffusion steps is 1000. Adam optimizer with a learning rate of 0.0002 is used to train the model. MoRAG-Diffuse was trained on an NVIDIA GeForce RTX 2080 Ti, with a batch size of 64, using initial weights of ReMoDiffuse\cite{zhang2023remodiffuse} for 50k steps.

\section{Prompt strategy}
Table ~\ref{tab:prompt_strategy} presents the LLM outputs for various text descriptions generated using our prompt function, demonstrating its effectiveness in motion retrieval. We observe that the generated descriptions are often correlated with other body parts, which led us to train part-specific retrieval models on full-body motion sequences.


\subsection{Spatial Composition}
Fig.~\ref{fig:spatial_comp} illustrates the composition workflow for combining part-specific motions, $R_{part}$, retrieved from corresponding part-specific databases, $\mathcal{D}_{part}$, using the LLM-generated part-specific descriptions, \texttt{text}$_{part}$, where $part \in \{ torso, hands, legs\}$.

\subsection{Significance of "position"}
\label{sec:position}
\begin{figure*}[ht]
  \centering
   \includegraphics[width=\linewidth]{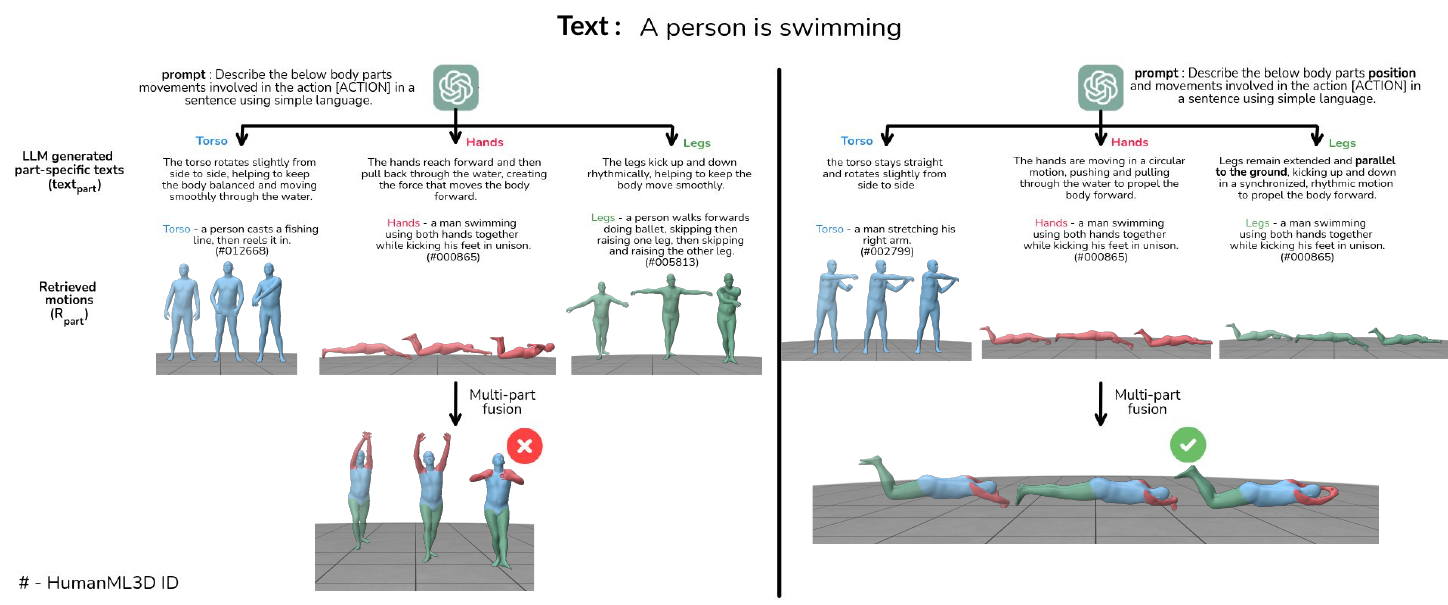}

   \caption{\textbf{Position Significance}: Impact of positional information in leg descriptions on motion retrieval accuracy and global orientation consistency in composed sequences for the text: \textit{"A person is swimming"}}
   \label{fig:position}
\end{figure*}

To train independent part-specific retrieval models, $MoRAG_p$, for $p \in \{torso, hands, legs\}$, it is essential to obtain movement information specific to each part. However, since the framework includes a composition step that combines the retrieved motions into a single motion sequence, relying solely on movement information is insufficient. The composition also requires positional information. For example, as shown in Fig.~\ref{fig:position} for the text \texttt{text} = \textit{"A person is swimming"}, explicitly prompting for the positional information allows the leg description to reflect its relative position to the ground, thereby retrieving the correct motion sample. The global orientation of the leg corresponding retrieved sample is important as it will determine the global orientation of the composed motion sequence.

\subsection{Issue with left/right retrieval strategy}
\label{sec:left-right}

\begin{figure*}[!t]
  \centering
   \includegraphics[width=\linewidth]{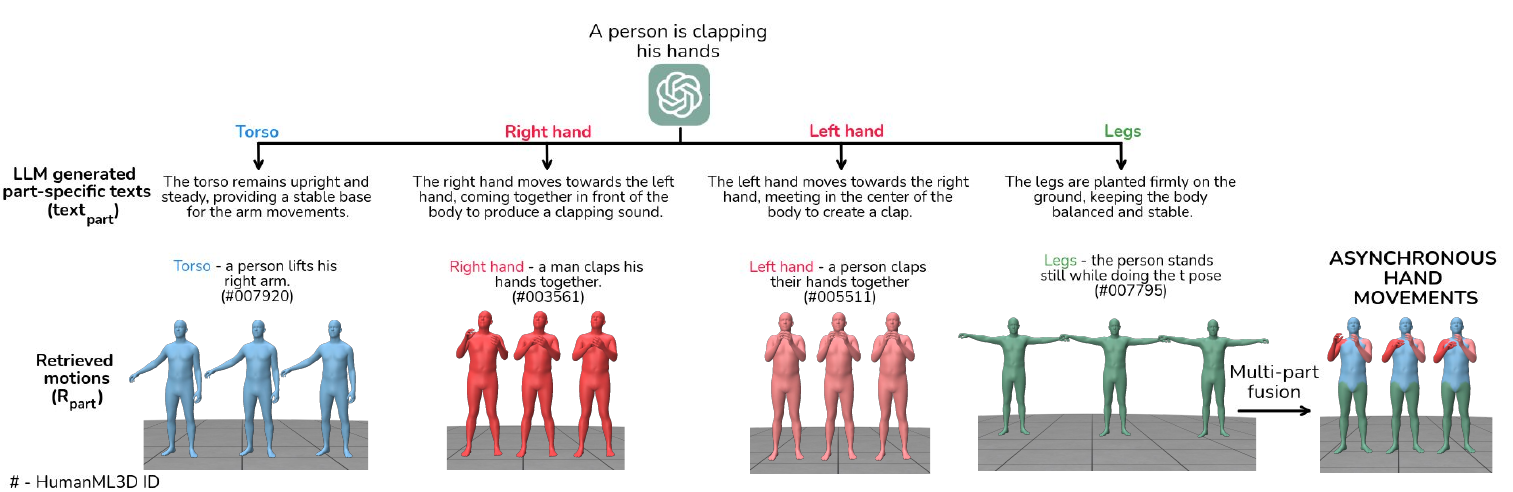}

   \caption{\textbf{Asynchronous motion caused by separate retrieval of left and right parts}: Illustration of asynchronous motion composition resulting from separate left and right hand retrievals for the text: \textit{"A person is clapping his hands."}}
   \label{fig:left-right}
\end{figure*}

In MoRAG, we avoided the strategy of retrieving the left and right hands and legs separately, as their movements can become asynchronous when composed into a single motion sequence. This issue is illustrated in Fig.~\ref{fig:left-right}, which demonstrates asynchronous motion composition for the input text \textit{"A person is clapping his hands"}.

\section{Qualitative Analysis}
We present qualitative analysis across three key aspects: (1) Generalizability, (2) Diversity, and (3) Zero-shot performance, for both MoRAG (retrieval model) and MoRAG-Diffuse (generative model). We also provide baseline comparisons for generalizability and zero-shot capabilities of MoRAG against TMR++\cite{lbensabath2024}, a state-of-the-art motion retrieval model, and ReMoDiffuse\cite{zhang2023remodiffuse}, which serves as the basis for MoRAG-Diffuse. Related videos can be found in project page.

\subsection{Generalizability}
The incorporation of part-specific descriptions \texttt{text}$_{torso}$, \texttt{text}$_{hands}$ and \texttt{text}$_{legs}$, generated by LLMs enables MoRAG to construct motion samples that accurately capture subtle variations in text, allowing for finer detail representation. By utilizing these detailed descriptions, the model retrieves more precise and nuanced motions, reflecting minor differences in body part movements or positions within the input text. This results in more realistic and contextually appropriate motion sequences. Example videos can be found in project page.

Conditioning these samples in the motion generation pipeline improves the model's comprehension of the language space, allowing it to effectively capture variations in text descriptions. This improvement is reflected in the Multi-Modal Distance metric. (Tab.~\ref{tab:metrics}) Example videos can be found in project page.

\subsection{Diversity}
MoRAG constructs diverse set of motion samples for a given input text \texttt{text} by utilizing both, (1) LLMs' ability to generate diverse text descriptions for a prompt and (2) various combinations of retrieved part-specific motion samples. Examples of these diverse samples can be found in project page.

The diverse samples produced by MoRAG improve the diversity of MoRAG-Diffuse, as indicated by the Diversity metric in Tab.~\ref{tab:metrics}. Related videos can be found in project page.

\subsection{Zero-Shot Performance}
Our approach facilitates the construction of motion sequences for unseen text phrases (zero-shot). This capability arises from two key aspects of the MoRAG framework: (1) it utilizes LLM-generated descriptions rather than the input text directly, and (2) it employs part-wise spatial composition. While the input text may be novel, the LLM-generated part-specific descriptions are not entirely unknown; relevant samples exist where the desired action is performed by specific body parts. Our method retrieves such samples independently for each body part and integrates them to construct motion sequences for unseen text descriptions. Example videos can be found at in project page.

Conditioning the samples constructed by MoRAG facilitates the generation of motion sequences for previously unseen text phrases in MoRAG-Diffuse. Example videos demonstrating this capability are available in project page.

\begin{table*}[!t]
\centering
\resizebox{\linewidth}{!}
{
\begin{tabular}{|c|ccc|ccc|}
\toprule
\textbf{\texttt{text}} & \multicolumn{3}{c}{\textbf{LLM Output}} & \multicolumn{3}{c}{\textbf{MoRAG part-specific retrieval}} \\
\midrule
 & \texttt{text}$_{torso}$ & \texttt{text}$_{hands}$ & \texttt{text}$_{legs}$ & \texttt{R}$_{torso}$ & \texttt{R}$_{hands}$ & \texttt{R}$_{legs}$ \\

\midrule


\shortstack{A person is standing\\and raising both hands} & \shortstack{The person's torso is upright\\ and still, while standing\\tall and balanced.} & \shortstack{The person's hands are being\\lifted up towards the sky, using\\their arms to extend upwards.} & \shortstack{The legs are steady and stable,\\acting as a strong foundation\\to support the body as the arms raise up.} & \shortstack{a person uses their\\hands to clap\\(\#002573)} & \shortstack{a person raises his\\hands above his head.\\(\#002315)} & \shortstack{a person raises his\\hands above his head.\\(\#002315)} \\
\midrule
\shortstack{A person is standing\\and raising single hand} & \shortstack{The person's torso is\\upright and still, while\\their arm raises up.} & \shortstack{One hand is lifted up from\\the side of the body,\\extended upwards and reaching\\toward the ceiling.} & \shortstack{The legs are supporting the\\person's weight, standing\\firmly on the ground.} & \shortstack{\\(person stands still and\\lifts right hand to\\face and mouth area\#000813)} & \shortstack{a person raises his\\right arm and then lowers it.\\(\#001179)} & \shortstack{a figure claps around\\shoulder height\\(\#007523)} \\
\midrule
\shortstack{A person is standing on\\one leg and raising\\both hands} &\shortstack{The person's upper body is\\straight and centered while\\standing on one leg. } & \shortstack{Both hands are lifted\\above the head, reaching\\towards the sky.} & \shortstack{One leg is holding the person's\\weight while the other is slightly\\lifted off the ground, balancing the body.}  & \shortstack{a person grabs their right\\foot and places it on\\their left thigh, and\\balances on one foot and\\then does the same\\with the other foot.\\(\#006123)} & \shortstack{raising and lowering arms.\\(\#011583)} & \shortstack{a person balances on\\their left leg and\\then their right.\\(\#009917)} \\
\midrule
\shortstack{A person is standing on\\one leg and raising\\single hand} &\shortstack{The person's body is\\upright and balanced on one\\leg, with the other leg\\lifted off the ground.} & \shortstack{One hand is raised up \\in the air, reaching toward\\the ceiling or sky.} & \shortstack{The standing leg is\\firmly planted on the ground,\\while the other leg is lifted\\up and may be slightly\\bent or straight.}  & \shortstack{the person raises their\\left foot up to their\\knee and then kicks their\\foot out, then returns\\their foot to their knee.\\(\#004012)} & \shortstack{a person raises his \\right arm and then\\lowers it.\\(\#001179)} & \shortstack{person balances on left\\leg then with arms up\\high has arms fully extended\\ keeping balance on left leg\\(\#004180)} \\
\midrule
\shortstack{A person is running with\\their arms crossed.} &\shortstack{The person's upper body is\\straight and upright, while\\their chest and shoulders\\may be slightly forward\\ as they run.} & \shortstack{The hands are crossed\\over the chest, alternating\\in front of the body\\as the person runs.} & \shortstack{The legs are moving back\\and forth in a rhythmic\\motion, propelling the\\ person forward as they run.}  & \shortstack{a person jogs forward\\with arms moving to\\his side.\\(\#002755)} & \shortstack{a figure stands in\\place, crossing its arms\\(\#003973)} & \shortstack{a person slowly jogs\\to the left to right\\and the jogs back into place\\(\#004273)} \\
\midrule
\shortstack{A person is running with\\their arms stretched out\\to the sides.} &\shortstack{The person's torso is upright\\and slightly leaning forward\\as they jog, with\\their chest and shoulders\\in a relaxed position.} & \shortstack{The person's hands are\\held out to the sides\\at shoulder level, moving\\rhythmically with each step.} & \shortstack{The legs are moving in\\a bouncing motion, alternating\\between the left and right\\sides as the person jogs forward.}  & \shortstack{jogging forward in medium pace.\\(\#000972)} & \shortstack{the person is flying like\\a airplane.\\(\#006433)} & \shortstack{a person slowly jogs\\to the left to right\\and the jogs back into place\\(\#004273)} \\
\midrule
\shortstack{A person is walking with\\their arms circling around.} &\shortstack{As the person runs,\\their torso is upright and\\ facing forward, with their\\chest and stomach comfortably relaxed.} & \shortstack{The person is moving their\\arms back and forth in a\\ swinging motion, with their\\arms stretched out to the sides.} & \shortstack{The legs are taking turns\\lifting off the ground and\\ propelling the person forward\\in a steady, rhythmic motion.}  & \shortstack{running from side to side.\\(\#014305)} & \shortstack{this person moves both\\arms out to his sides in\\a large circular motion\\then walks forward.\\(\#005375)} & \shortstack{a person doing a\\casual walk\\(\#004183)} \\
\midrule
\shortstack{A person is walking with\\their arms crossed.} &\shortstack{The person's upper body\\remains upright while they\\walk with their arms crossed,\\with their spine and shoulders\\in a stable position.} & \shortstack{The person's hands are placed\\on opposite shoulders,\\crossing in front of their\\torso as they walk.} & \shortstack{The legs are alternately\\lifting and stepping forward,\\propelling the person forward\\while they walk.}  & \shortstack{a person walks on uneven\\ground whilst holding\\on to handrail.\\(\#001029)} & \shortstack{a person crosses their\\arms in front of their\\chest, then drops them\\back at their sides.\\(\#001014)} & \shortstack{a person walks forward\\while holding out their\\arms for balance\\(\#002087)} \\
\midrule
\bottomrule
\end{tabular}
}
\captionof{table}{\textbf{Prompt Examples :} We illustrate the LLM-generated part-specific outputs for text descriptions alongside their corresponding top-1 retrieval results to demonstrate the effectiveness of our prompt strategy. The HumanML3D ~\cite{Guo_2022_CVPR} ID for the retrieved motions is indicated with the \# symbol.} 
\label{tab:prompt_strategy}
\end{table*}

\section{Metrics}
\label{sec:metrics_explanation}

For quantitative evaluations, we adopt the performance metrics used in ReMoDiffuse\cite{zhang2023remodiffuse}, which include R Precision, Frechet Inception Distance (FID), Multi-Modal Distance, Diversity and Multimodality. For R Precision and MultiModality, higher scores indicate superior performance. Conversely, lower scores are preferred for FID and Multi-Modal Distance. For Diversity, performance improves as the score more closely aligns with the real motions.

(1) R Precision evaluates how well the generated motion sequences semantically match the text descriptions. We calculate Top-1, Top-2, and Top-3 accuracy by measuring the Euclidean distance between each motion sequence and 32 text descriptions (one ground truth and 31 randomly selected descriptions). (2) FID calculates the distance between features extracted from real and generated motion sequences. (3) Multi-modal distance (MM Dist for short)
measures the average Euclidean distance between the feature vectors of text and generated motions. (4) Diversity measures the variability and richness of the generated sequences. Average euclidean distance is calculated between the two-equal sized subsets which are randomly sampled from the generated motions from all test texts. (5) Multimodality measures the diversity of generated motions for a given text. We generate 10 pairs of motions for each text and compute average distance between the each pair feature vectors. 

\subsection{Challenges in applying motion retrieval metrics to spatially composed sequences}
To evaluate text-to-motion retrieval methods such as TMR\cite{petrovich23tmr} and TMR++\cite{lbensabath2024}, the similarity between the text corresponding to the retrieved motion sample and the input text is computed. However, for our approach, which involves the spatial composition of motion sequences, there is no single corresponding text for the entire composed sequence. As a result, these metrics cannot be computed in our case.

\newpage
    
    
{\small
\bibliographystyle{ieee_fullname}
\bibliography{egbib}
}

\end{document}